\documentclass{article}

\usepackage{lscape}

\usepackage{amsmath,amssymb,amsfonts}
\usepackage{algorithmic}
\usepackage{graphicx}
\usepackage{textcomp}

\usepackage[sorting=none]{biblatex}
\usepackage{booktabs}
\usepackage{multirow}
\usepackage{comment}

\def\BibTeX{{\rm B\kern-.05em{\sc i\kern-.025em b}\kern-.08em
    T\kern-.1667em\lower.7ex\hbox{E}\kern-.125emX}}

\DeclareTextFontCommand{\textbf}{\bfseries}
\DeclareTextFontCommand{\textit}{\itshape}

\begin{document}

\begin{center}

{\Large \bf A Review of Multi-Modal Large \\Language and Vision Models\footnote{Acknowledgement: This work was part-funded by Science Foundation Ireland through the Insight the SFI Research Centre for Data Analytics (SFI/12/RC/2289\_P2).}}

\vspace{1cm}

{\large \bf Kilian Carolan$^{(1)}$, Laura Fennelly$^{(1)}$ \\and Alan F. Smeaton$^{(1,2)}$}
\end{center}
\vspace{1cm}

\noindent 
{[1] School of Computing, Dublin City University, Glasnevin, Dublin 9, Ireland.}\\
{[2] Insight SFI Centre for Data Analytics, Dublin City University, Glasnevin, Dublin 9, Ireland.}\\

\begin{abstract}

Large Language Models (LLMs) have recently emerged as a focal point of research and application, driven by their unprecedented ability to understand and generate text with human-like quality.
Even more recently, LLMs have been extended into multi-modal large language models (MM-LLMs) which extends their capabilities to deal with image, video and audio information, in addition to text. This opens up applications like text-to-video generation, image captioning, text-to-speech, and more and is achieved
either by retro-fitting an LLM with multi-modal capabilities, or building a MM-LLM from scratch.  
This paper provides an extensive review of the current state of those LLMs with multi-modal capabilities as well as the very recent MM-LLMs. It covers the historical development of LLMs especially the advances enabled  by transformer-based architectures like OpenAI's GPT series and Google's BERT, as well as the role of attention mechanisms in enhancing model performance.
The paper includes coverage of the major and most important of the LLMs and MM-LLMs and also covers the techniques of model tuning, including fine-tuning and prompt engineering, which tailor pre-trained models to specific tasks or domains.
Ethical considerations and challenges, such as data bias and model misuse, are also analysed to underscore the importance of responsible AI development and deployment. Finally, we discuss the implications of open-source versus proprietary models in AI research. Through this review, we  provide insights into the transformative potential of MM-LLMs in various applications.

\end{abstract}



\section{Introduction}

Large Language Models (LLMs) are one of the hottest topics in artificial intelligence (AI) research and interest in them and how they can be used in generative AI applications has spilled into the mainstream media.
Their popularity in the public sphere has been triggered by the arrival of ChatGPT, first released to the public in 2022 \cite{ray2023chatgpt}.
By LLMs we refer to language models based on the Transformer architecture, with the notable example of `Generative Pre-trained Transformers (GPT)' from OpenAI first introduced as GPT-1 in 2018~\cite{radford2018improving}. The sudden popularity of LLMs arises because of their demonstrated usefulness in supporting a wide range of applications and tasks including text summarisation~\cite{10.1145/3649506}, text-to-image~\cite{10.1145/3581783.3612012} and text-to-video~\cite{lian2024llmgrounded} generation, conversational search~\cite{AI202380}, machine translation~\cite{moslem-etal-2023-adaptive} as well as their role in many Generative AI (GenAI) applications. That role in GenAI is included in a comprehensive review of the literature consisting of over 1,300 articles in~\cite{GUPTA2024100066}.

Apart from the GPT-n family of LLMs from OpenAI, other notable examples of proprietary LLM's which are in the public eye include  Google's Gemini/BARD~\cite{team2023gemini} and Anthropic's CLAUDE~\cite{kane23} while the most well-known open source LLMs are Meta's LLaMA~\cite{touvron2023llama}, Google's PaLM and Falcon from the UAE's Technology Innovation Institute.
The release of a new LLM or the announcement about an update to an existing one can not only pique the interests of the research community but often also attracts media attention. This makes keeping abreast of all the LLMs that are available, how they compare against each other and what they are being used for, a challenging task.

In this review, we assess the current state of LLM's but with a specific focus on visual/multi-modal LLM's and on their technical aspects, how they can be optimised for a specific task. In~\cite{10433480} the authors presented a short, thorough review of LLMs covering  their history, architectures, training methods, applications, impacts, challenges and significance but their scope did not cover models that have been trained on, and can generate output which crosses the boundaries of different modalities namely  text, images, videos, and audio.
Our review is complementary to that in~\cite{10433480} and guided by the evaluation of LLMs and MM-LLMs, considers factors such as open-source versus proprietary owned models, the benefits of choosing open-source and the computational costs and resources required to pursue open-source. We also address issues such as whether the fine-tuning methods or architectural components in each model that should be considered in order to minimise cost and what evaluation methods are used to assess the quality of these models.

Many ethical concerns have been raised in the literature concerning the use of LLMs, the data used to train them~\cite{head2023large,tokayev2023ethical}, the energy costs associated with their creation~\cite{samsi2023words,rillig2023risks} and the fact that the largest and most useful of these models are owned by a small number of big tech companies.  This review looks to address if and how a Multi-Modal LLM, in particular an open source MM-LLM can be used in practical multimedia applications.  We examine how they are trained to become MM-LLMs and how the visual components which defines them as Multi-Modal were incorporated to their base LLM. 
Additionally, we highlight the comparison between  closed source vs. open source from an ethical standpoint, data control issue and application use case.

Other aspects of such models, of which there are several, fall outside the scope of this review.

The rest of this paper is organised as follows. In the next section we briefly re-cap on  large language models to explain what they are, the history of their development and the importance of the attention mechanism.  We then examine look at the pros and cons for using proprietary vs. open source LLMs.
This is followed by a review of the largest and most popular of the mostly text-based LLMs, covering the
LLaMA family, Mistral-7B, Falcon and its variants 
Following that we review the major vision models and the Multi-Modal Large Language Models (MM-LLMs) BLIP-2, CLIP, LLaVA, Kosmos-1, MiniGPT4, and MPLUG-Owl. This is followed by a review of ways to modify a foundational model for specific applications including model fine-tuning, both full tine-tuning and parameter efficient fine-tuning covering Low Rank Adaption (LoRA), Quantised Low-Rank Adaption (QLoRA) and Supervised Fine-Tuning (SFT), prompt engineering and Reinforced Learning Human Feedback (RLHF).  Hallucinations, which refer to outputs from a LLM or a MM-LLM which are obviously wrong, factually incorrect, or unrelated to the input prompt, are then presented along with approaches to mitigate their impact, especially for MM-LLMs. Finally, 
model evaluation and benchmarking to
assess the performance of both pre-trained and fine-tuned LLMs and MM-LLMs is an important consideration in any use of such models and a number of commonly used benchmarks for assessing the performance of an LLM in common-sense reasoning tasks, are presented.


\section{What is a Language Model?}

\subsection{The History of Language Models}

Language Models (LMs) are an important technology in Natural Language Processing (NLP). Using statistical methods,  
LMs were machine-learning based, trained to predict the next word in a sequence by analysing a large amount of text corpora to understand and replicate the patterns of language use, styles, and the nuances of human language~\cite{techtargetWhatLanguage,tdsBriefHistory}.

The development of LMs into LLMs can be traced through the historical advances in NLP. The first approaches to NLP for applications like machine translation and speech recognition were characterised by rule-based approaches, which  evolved into the uses of statistical approaches such as Hidden Markov and N-gram models~\cite{anandika2021review}. These statistical approaches  analysed the sequences, order, and frequency of words in text corpora, but were limited in their ability to capture context or semantics of the data efficiently. While Neural Networks (NNs) were initially discovered in the 1950s, their use in NLP tasks came about much later due to their computational requirements~\cite{castilho2017neural}. However, in the 2010s, a new breakthrough was marked in LMs and NLP tasks with the introduction of Word Embedding models such as Word2Vec and GloVe~\cite{patil2023survey}. Word embedding is a technique in which text data can be represented as continuous vectors by creating a virtual embedded space where
semantic relationships between words can be captured and understood. The innovation of word embedding laid down the initial works for the resurgence in deep learning. 

LM's next advancement was seen with the utilisation of Recurrent Neural Networks (RNNs) and Long Short-Term Memory (LSTM). The incorporation of these deep learning methods into LMs marked a significant leap forward in enhancing  contextual awareness in language processing~\cite{cui2023survey,tdsBriefHistory,gopenaiLargeLanguage}.
Models like RNNs enabled the processing of a whole input sequence, one word after the other, which could then be used to predict the next word in the sequence. This is unlike N-gram models which could only capture the context of words that were close by and failed at capturing the context of words that were far from each other. There were still limitations and challenges with RNN and LSTM models, specifically with capturing long-range dependencies in text data. As  input sequences grew longer, it created a bottleneck for all the relevant information to pass through and for  context to be captured as sequences could not be trained in parallel~\cite{tdsChoosingNeural}.

In 2017 a pivotal paper was published~\cite{vaswani2023attention} which led to what have become known as LLMs. This paper introduced the Transformer architecture and one of its innovations was that it used attention mechanisms to enable parallel processing of text sequences. This removed the limitation found in RNN and LSTM models associated with long-range dependencies in text. It also allowed for LMs to pay attention to, and to analyse different parts of the input, so that the whole input is understood as is its relationships and the importance of each part~\cite{vaswani2023attention,huang2023language,cui2023survey}. 

In 2018 researchers in Google  and in OpenAI each introduced a language model based on the Transformer architecture. Google announced the Bidirectional Encoder Representations from Transformers (BERT)~\cite{devlin2018bert} and OpenAI released their first Generative Pre-trained Transformers (GPT).  These Transformer-based language models along with the vast amounts of text corpus and increased computational power available to train them, has led to development of LMs into LLMs~\cite{gopenaiLargeLanguage,radford2018improving, devlin2018bert}. 

However, the surge in popularity for the use of, and research into LLMs was not until the arrival of OpenAI’s ChatGPT which was first released to the public in November 2022. This demonstrated the ability of LLMs to ``understand'' and generate human-like language in conversational search, document summarisation and other generative AI applications~\cite{forbesShortHistory}.

\subsection{Attention Mechanisms}

The Transformer architecture has been a paradigm shift in the field of NLP. First introduced in~\cite{vaswani2023attention}, it is an architecture that uses only attention mechanisms, specifically Self-Attention and Multi-Headed attention. Recently, methods utilising the attention mechanism  have dominated this area. 

Attention mechanisms allow a model to focus on important parts of an input sequence without being affected or reliant on dependencies of distance in the sequence. The self-attention mechanism relates to different positions of a single sequence in order to compute a representation of the sequence. Each input sequence is broken down into smaller, unique, linear query, key and value vectors for better context understanding. 

The Multi-Head attention (MHA) mechanisms repeats the self-attention computation multiple times in parallel. MHA uses multiple heads to attend to all the different aspects of an input sequence simultaneously. This allows it to provide high-quality results. However, this can be computationally expensive due to the parallel processing and can potentially lead to performance deterioration. 

Multi Query Attention (MQA) shares one key head and a single value head across all unique query heads. As the number of heads for key and value are reduced, the GPU memory required for storage is also reduced and allows for faster processing.   GPU memory saving can then be utilised to increase batch size, and therefore increase overall efficiency. The disadvantage to using MQA over MHA is that the former only uses one head for a key-value pair therefore it is prone to missing potentially important aspects of the input sequence compared to the more in-depth analysis of MHA. 

\begin{figure}[!ht] 
    \begin{center} 
        \includegraphics[width=0.9\textwidth]{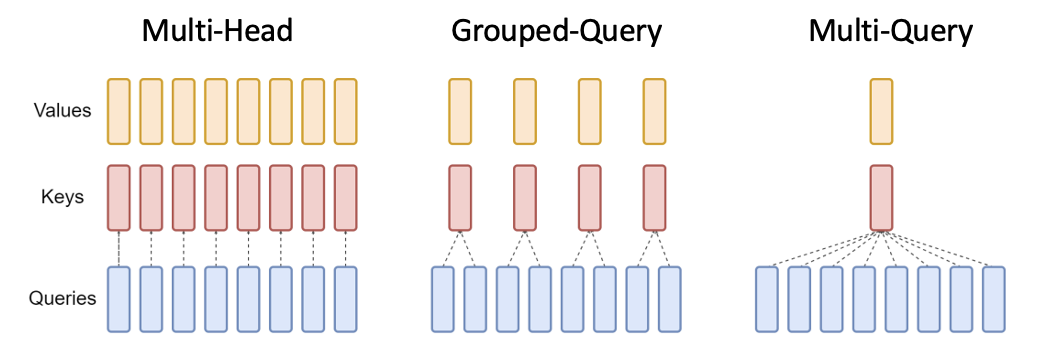}
    \end{center}
    \label{fig:attentionma}
    \caption{A summary of how an input sequence is decomposed into query, key, and value vectors across the various attention mechanisms, taken from~\cite{ainslie2023gqa}.}
\end{figure}

The compromise between MHA and MQA is called Grouped-Query attention (GQA) and  3 variations are shown in Figure~\ref{fig:attentionma}. GQA allows for faster processing than MHA while retaining more detail than MQA through attention within groups. This allows the model to effectively comprehend longer input sequences. GQA addresses this issue found in MHA by assigning each key-value head a corresponding query group, rather than a unique linear query. This optimises performance while mitigating risk of performance degradation~\cite{visoLlamaNext, mediumNavigatingAttention,ainslie2023gqa}.

\section{Proprietary vs. Open Source LLMs}

Throughout 2023, the ongoing surge in research and development of LLMs continued at pace. Prominent companies such as OpenAI and Google competed to develop an LLM that would dominate the market. It was previously considered that the larger the LLM, the greater the competitive advantage that would be obtained. The cost for researchers to build an open sourced LLM had required substantially large funding, as the creation of LLMs demanded extensive data and GPU resources, which could cost anywhere from €1 to €100 million. 
This was until Meta released their open sourced LLM called LLaMA, their belief being that to foster innovation, enhance security, and improve safety in LLMs, it was important to release an open sourced LLM~\cite{linkedinWhatLearned}.
Some open source models like Meta’s LLaMA-2 and Google’s PaLM 2, are now available to use free of charge as opposed to proprietary LLMs such as OpenAI‘s GPT or Google’s BARD which charge enterprises based on their usage of the model~\cite{digidayCaseAgainst,ibmOpenSource}. 

It should be noted that Google has acknowledge the limitations of proprietary LLMs and that there would only be limited time before their own model would be replicated or surpassed by open source LLMs. This is because the innovations that were driving the successes in the open source community had addressed challenges that Google had already been grappling with~\cite{linkedinFutureLLMs, semianalysisNoMoat}.

Open source LLMs offer several advantages for researchers and for entrepreneurs, primarily due to their long-term cost-effectiveness but also because they provide transparency and flexibility in terms of their methodologies, architecture, and training data. The visibility in open source models aids with audits while facilitating adherence to both ethical and legal standards. This will be especially important when the European Union AI act comes into force in 2025 where such openness and transparency about training data will be a requirement if companies are to deploy use of their LLMs within the European Union~\cite{EUAI}.  Another benefit is that open source grants researchers and developers’ complete control over their own data which they may wish to combine with, or to fine-tune, an LLM. Any sensitive data that might be used will remain within their system which aids in reducing the risk of data leaks or unauthorised access. Furthermore, efficient optimisation of open source LLMs can result in reduced latency and improved performance. 

Open source LLMs have been utilised across a diverse range of industries, including healthcare, law, and finance among others. However, there are noteworthy concerns associated with the use of open source LLMs that should be considered. Firstly, there is a lack of formal service agreements between the provider of an open source LLM and the entities utilising it for development. Additionally, there is no guarantee that a company or developer will continue to offer support or continued development for an open source LLM in the future. Furthermore, as open source communities drive technological advancements, proprietary-owned LLMs may progress to become a more practical and reliable option in the future. Lastly, it is important to consider that not all open source LLM adhere strictly to the open source philosophy. For instance, Meta’s LLaMA-2 has imposed certain usage conditions through its acceptable use policy~\cite{digidayCaseAgainst, ibmOpenSource, OpenSource}.

\section{Specific Large Language Models}
\label{sec:LLMs}

In this section we provide a summary review of the most popular and important of the LLMs which originate in the text domain, some of which are mentioned in~\cite{10433480}. While primarily aimed at generating text as output,  some
of these have a multi-modal component which has been retrofitted afterwards,

\subsection{GPT}
GPT-n, first introduced as the GPT model refers to a family of LLMs developed by OpenAI, which at the time of writing now includes a MM-LLM named GPT-4. The first model in this family was GPT-1, where GPT means Generative Pre-Training, and was trained on a large, diverse corpus of text using a 12-layer decoder-only transformer with masked self-attention heads. It was shown to improve performance of the state-of-the-art NLP tasks on 8 of 12 datasets studied~\cite{radford2018improving}.

GPT-3 was introduced in 2020 and shares a relatively similar architecture to GPT-1 and GPT-2 but was scaled up in terms of model size, training time, and training data. GPT-2 first showed that language models can demonstrate impressive NLP performance in areas like  text comprehension and summarisation, unsupervised when trained on larger and larger datasets. GPT-2 was a 1.5 billion parameter model and GPT-3 was a 175 billion parameter model~\cite{brown2020language},\cite{radford2019language}.

GPT-3.5 and 4 are of particular importance in the development of LLMs with GPT-3.5 being the backbone of ChatGPT, released to the public in November 2022 and now synonymous with generative AI.
GPT-4 is the first of the OpenAI GPT models to have multi-modal capabilities. This model is able to accept images and text and to produce text and image outputs. While the ability to  take images as prompts is an important development, GPT-4 also shows a significant improvement on NLP tasks  when compared to its predecessors, scoring in the 90\% percentile in the bar exam when compared to humans while its immediate predecessor, GPT-3.5,  scored in the bottom 10\%. However, the model suffers from the same issues as its predecessors with hallucinations, limited context window size, and an inability to learn incrementally including from from experience. However what is known about GPT-4 is that it is a transformer-based model trained on a combination of public data from the internet and  licensed content from third parties.
The model is then fine-tuned using Reinforcement Learning from Human Feedback (RLHF)~\cite{achiam2023gpt}.

\subsection{Claude}

The Claude family of large language models developed by Antrhophic and first introduced as Claude 1 in March 2024, is a LLM designed to help with NLP tasks such as summarization and question-answering as well as with coding. It was introduced with claims that Claude was less likely to produce harmful outputs, and could take better direction and tone and behaviour than other models~\cite{introClaude2023}.

Claude 3 was introduced in 2023 and is a suite of models ranging in size and complexity, coming in three flavours: Opus, Sonnet, and Haiku. While specific details of the model architecture are not available, Claude 3 has multi-modal capabilities allowing for the processing of visual data, it claims excellence on visual queries based on images given as prompts. It also has what is possibly the largest context window of 200,000 tokens, which is the size of the input that an LLM can consider in one go when making predictions or generating text.
Claude 3 is trained with a mixture of public and private data as well as with synthetic data  With the public data crawled with a knowledge endpoint of August 2023. 
The claimed benchmark performance reported by Antropic the  developers of Claude, makes it on par with or better than the other large LLMs but   there is still a lack of clarity on specific architecture details.

\subsection{Gemini}
Gemini introduced by Google in 2023, is another ``family'' of, rather than a single MM-LLM which includes capabilities for text, audio, images and video. It boasts state-of-art results on multiple benchmarks at time of release. Of particular interest is the Gemini Ultras achievement on image and text benchmarks where for example it scored 64\% in the MMMU benchmark~\cite{yue2023mmmu} containing images on multi-discipline tasks requiring college-level subject knowledge, outperforming the previous best model by more than 5 percentage points.

Gemini 1 is available in three different flavours, Ultra, Pro and Nano referring to the size of the models. Each model is trained using the Transformer architecture utilising the MQA mechanism and can accept a context length of 32k tokens. As with GPT, the Gemini family would appear to follow the reasoning of ``bigger is better'', and have addressed shortcomings of Google's previous flagship models by adding more hardware. A key difference between GPT-4 and Gemini is the fact that the model can also output images.

\subsection{LLaMA}

LLaMA is a series of open source LLMs developed by Meta AI. The goal for Meta was to provide an accessible and open sourced LLM to help make advancements in research and application development in the sub field of NLP, thus enabling others in the research   community to access the vast infrastructure resources to study and explore the area of LLMs' rapidly evolving fields of use. 
LLaMA comprises a collection of foundation language models with parameter sizes ranging from 7B to 65B. Unlike other models that prioritise training speed, LLaMA was designed to run on a single GPU, with the objective being to focus on the inference speed rather than the speed of the fastest model to train~\cite{touvron2023llama, metaIntroducingLLaMA}.

LLMs were based on the common belief that more parameters will lead to better performance and so the bigger the model, the better the performance. A recent study by Hoffman {\em et al}.~\cite{rae2022scaling} suggested that smaller models trained on more data can yield better performance and results for a given computational budget. Meta AI trained the LLaMA language models using publicly available datasets exclusively, avoiding the use of proprietary or inaccessible data. The main objective of LLaMA was to achieve the best possible performance at different computational budgets for inference. Despite its smaller size, LLaMA-13B outperforms GPT-3 on most benchmarks, while LLaMA-65B is competitive with other LLMs like Chinchilla or PaLM-540B~\cite{touvron2023llama, mediumReviewLLaMAOpen, hoffmann2022training}.

LLaMA model architecture is based on the original Transformer architecture as described in~\cite{vaswani2023attention}. Additional improvements were made to the model by incorporating methods proposed by other LLMs. To make the training stable, LLaMA uses a technique called pre-normalisation, as demonstrated in the GPT-3 model. This method normalises the input of the transformer, rather than normalising the output, using the RMSNorm function. This helps to reduce computational time and complexity. LLaMA improves performance through utilising an activation function called SwiGLU. LLaMA uses Rotary positional embedding (RoPE) a technique that rotates the token embedding in a high-dimensional space. RoPE preserves the original information, while adding positional understanding. This can save memory and computational resources, making it a more efficient choice for large-scale models~\cite{touvron2023llama, UnderstandingRotary, mediumReviewLLaMAOpen}.

LLaMA was compared with non open source models such as PaLM, GPT-3 and Chinchilla using a total of 20 benchmarks to evaluate Zero-shot and Few-shot learning. Despite LLaMA being a smaller model, it outperformed or stayed competitive with PaLM, GPT-3 and Chinchilla in areas such as Common Sense Reasoning, Closed-book Question Answering, Reading Comprehension, and Code Generation. 
LLaMA did not perform as well as PaLM and Chinchilla in Massive Multi-task Language Understanding (MMLU).  However, it was found that small amounts of instruction fine-tuning applied to the model improved the performance for MMLU. LLaMA-65B efficiency was measured against areas such as truthfulness, bias, and toxic language. The following benchmarks were used: RealToxicityPrompts, CrowS-Pairs, WinoGender, and TruthfulQA. These are covered in more detail in Section~\ref{sec:benchmarks} and in~\cite{touvron2023llama, UnderstandingRotary, mediumReviewLLaMAOpen}.

\subsection{LLaMA-2 and  LLaMA-2 Chat}

In July 2023, Meta AI released LLaMA-2 and LLaMA-2 Chat, an updated version of LLaMA. LLaMA-2 and LLaMA-2 Chat are a collection of models ranging from 7B to 70B parameters in size. They key enhancements in LLaMA-2 included a 40\% increase in the size of the pre-training corpus and doubling the context length from 2048 to 4096 tokens. One notable difference in LLaMA-2 and LLaMA-2 Chat from their predecessor LLaMA, is the adoption of the fine-tuning method Reinforcement Learning from Human Feedback (RLHF) in both  LLaMA-2 and LLaMA-2 Chat. This fine-tuning technique is discussed in more detail in Section~\ref{subsec:FT}.
Although Meta AI have continued to use publicly available data for training, LLaMA-2 incorporates a new mix of data and have implemented increased safety measures to ensure the model utilised safety-specific data. While LLaMA was released as an open-sourced model with non-commercial license, LLaMA-2 introduces a commercial license to encourage collaborations and facilitate the exploration of new applications. Additionally, Meta AI provided the weights and initial code used to help simplify the process for developers and researchers to build upon existing work. LLaMA-2 adopts a similar architecture to LLaMA, with an important enhancement, the integration GQA mechanism within its transformer-based framework~\cite{touvron2023llama2, ankursnewsletterLLaMALLaMA, GameChanger2024}.

In terms of performance tasks and benchmarks utilised for LLaMA-2 and LLaMA-2 Chat, similar ones were used as in the case of LLaMA. LLaMA-2 generally out-performed most other open sourced LLMs across various performance tasks, with the exception of coding-based tasks. 
However, when compared to proprietary LLMs such as GPT-4 and PaLM-2, LLaMA-2 presented lower performance. It closely aligned with results of GPT-3.5 and PaLM in terms of performance outcomes~\cite{touvron2023llama2, kluWhatGrouped, visoLlamaNext}.

\subsection{MedAlpaca}

In October 2023 MedAlpaca, a collection of LLaMA models fine-tuned for biomedical tasks using open-sourced biomedical datasets was released. The objective of MedAlpaca was to address the need for open-source LLMs  that could be deployed on-premises to uphold privacy for personal data, specifically in the medical domain where the safeguard of patient privacy is of important concern. MedAlpaca utilised Low-Rank Adaption (LoRA) and Supervised Fine-Tuning (SFT), these fine-tuning methods are covered in more detail in Section~\ref{subsec:FT} and in~\cite{mlopshowtoParameterEfficientFineTuning,hu2021lora}.

The model was  evaluated by its performance on the United States Medical Licensing Examination (USMLE), a standardised assessment taken by medical students to qualify as physicians.  Performance was evaluated on zero-shot prompts and interestingly, MedAlpaca 13B  achieved better accuracies compared to LLaMA 13B, with 47.3\%, 47.7\%, and 60.2\%, in steps 1,2 and 3 of the USMLE assessment. However, when employing LoRA fine-tuning for MedAlpaca 13B LoRA, the performance decreased to 25.0\%, 25.5\%, and 25.5\%. These results indicate that the implementation of LoRA, while computationally efficient, was less optimal compared to using just SFT~\cite{ han2023medalpaca}.

\subsection{Mistral 7B}
Mistral 7B is a 7-billion parameter language model, claiming superior performance and efficiency when compared to  larger models such as LLaMA (Version 1 and 2, 34B and 13B parameters respectively). The developers of Mistral 7B highlight how in particular, a chat model built on Mistral 7B outperforms the LLaMA-2 13B Chat model. Mistral 7B attributes GQA mechanisms, similar to LLaMA-2,  and Sliding Window Attention (SWA) as reasons for its performance and efficiency. 
SQA, introduced as part of Long-former architecture~\cite{child2019generating}, allows for more efficient processing of longer text sequences. A key component of this is having multiple stacked transformers, comparable to CNN convolutional layers. 

\subsection{Falcon-7B, 40B}

In May 2023, Abu Dhabi’s Technology Innovation Institute (TII) announced Falcon, a series of open sourced LLMs comprised of Falcon-7B, Falcon-40B, Falcon-7B-Instruct and Falcon-40B-Instruct. All  models were released under the Apache 2.0 license. Falcon-7B-Instruct and Falcon-40B-Instruct are specialised versions of Falcon-7B and Falcon-40B models, which have been fine-tuned on instruction and conversational data so that they can be leveraged for chatbot style tasks. Falcon enables the use of their LLMs for unrestricted commercial purposes without any usage costs or royalties, facilitating widespread adoption. This means the LLMs can be fine-tuned with various datasets for different application purposes. 

The Falcon LLM team at TII also developed a high-quality pre-training dataset called “The RefinedWeb Dataset”, which they released as an extract of 600 billion tokens for the open-source community to use in their own LLM endeavours~\cite{tiiUAEsTechnology,  ExploringFalcon,huggingfaceFalconLanded}.

The architecture of the Falcon models is transformer-based, incorporating the use of the MQA mechanism for efficient handling of large-scale tasks, allowing for enhanced scalability of inference by reducing memory consumption and enabling faster processing. Similar to LLaMA, Falcon utilises RoPE to maintain positional understanding and minimise computational resources. Another notable feature in Falcon models is Flash Attention, an algorithm that optimises for speed and memory-efficiency in transformers which is achieved through tiling and re-computation to minimise the number of memory read/writes between levels of GPU high bandwidth memory (HBM) and GPU. This results in faster training of transformers and enables longer context, leading to higher quality models and improved performance across various tasks. Unlike LLaMA, Falcon did not employ the SwiGLU activation function to enhance performance due to its additional memory cost. After reviewing the use of GLU activations for scaling, it was decided that the extra memory cost of GLU was not worth it for cost-efficient training~\cite{almazrouei2023falcon,ExploringFalcon,huggingfaceFalconLanded}.

The Falcon models have been trained on 1.5 trillion tokens, and their performance is believed to be heavily influenced by the carefully curated nature of the pre-training data used. LLMs are affected by the data they are trained on, and their sensitivity varies with changing data. What distinguishes Falcon from other LLMs is  the pre-training data which is predominantly based on the RefinedWed dataset. The Falcon LLM team created the RefinedWed dataset, which is a massive web dataset based on CommonCrawl. TII focused on scaling and improving the quality of web data by employing large-scale de-duplication and strict filtering, resulting in high quality pre-training data. The creation of the RefinedWeb dataset for the Falcon LLM demonstrated that the use of high quality pre-training data can lead to powerful results~\cite{almazrouei2023falcon, huggingfaceFalconLanded, penedo2023refinedweb}.

\subsection{Falcon-180B}

In September 2023 TII introduced Falcon-180B, a big advancement in the Falcon series and trained on 3.5 trillion tokens from the RefinedWeb dataset, over double what was used for previous models. Additionally, TII, released Falcon-180B-chat, a variant fine-tuned specifically on instruction and conversational data. With its 180 billion parameters, the model proved promising performances with comparable results to leading models such as PaLM 2-Large, GPT-3.5, and GPT-4. In term of scale, Falcon-180B is 2.5 times larger than LLaMA-2. However, a notable drawback of this large open-sourced model is its substantial memory requirement. It is reported that a minimum of 320GB of memory is required for optimal operation of Falcon-180B, whereas Falcon-40B requires 40GB of memory, and Falcon-7B requires 15GB making it accessible to the open source community for inference and fine-tuning on modest hardware. The imposed significant burden of Falcon-180B on resources contradicts one of the major advantages of open-source models, having the ability to run open-sourced models on consumer level hardware which allows for accessibility ~\cite{almazrouei2023falcon, paperspaceIntroducingFalcon}.

The performance tasks and benchmarks utilised for evaluating the Falcon Series of models included common sense reasoning benchmarks such as HellaSwag, Winogrande, AI2 Reasoning Challenge (ARC), MMLU and OpenBookQA. Other additional benchmarks used included PIQA (Physical Interaction: Question Answering) and BoolQ. These performance tasks and benchmarks are discussed in more detail in Section~\ref{sec:benchmarks} Model Evaluation and Benchmarking
 and in~\cite{almazrouei2023falcon, paperspaceIntroducingFalcon}. 
 
\subsection{Grok-1}

The final LLM to be included in this review is Grok-1, developed by xAI and released by OpenAI in March 2024.
Grok-1 is xAI's frontier LLM with a claimed size of 314B parameters. Its architecture is like most of the other LLMs and is an autoregressive Transformer-based model but with a mixture of 8 experts.
It achieves 63.2\% on the HumanEval coding task, a Python coding completion task, and 73\% on MMLU.  On these and other benchmarks it does not achieve the performance of models with larger amounts of training data such as GPT-4 or Claude 2 but it does out-perform models trained on similar sized datasets.

A summary of some of the features of the LLMs covered here is presented in Table~\ref{tab:LLM-comp}.

\begin{landscape}
    
\begin{table*}[htb]
\centering 
\begin{tabular}{l|llllllll}
\toprule
\multirow{2}{*}{Model} & No. & Commercial  & \multirow{2}{*}{License} & Attention  & Pre-training  &  VRAM or & Open & Fine \\
& Parameters & Use && Mechanism & Token Length & RAM Required & Source & Tuneable \\
\midrule
LLaMA & 7 Billion & No & LLaMA License & MHA & 1 Trillion & 6GB VRAM & Yes & Yes \\
LLaMA & 13 Billion & No & LLaMA License & MHA & 1.5 Trillion & 10GB VRAM & Yes & Yes \\
LLaMA & 65 Billion & No & LLaMA License & MHA & 1.5 Trillion & 40GB VRAM & Yes & Yes \\
\midrule
LLaMA-2 & 7 Billion & Yes & LLaMA-2 License & GQA & 2 Trillion & 6GB VRAM & Yes & Yes \\
LLaMA-2 & 13 Billion & Yes & LLaMA-2 License & GQA & 2 Trillion & 10GB VRAM & Yes & Yes \\
LLaMA-2 & 70 Billion & Yes & LLaMA-2 License & GQA & 2 Trillion & 40GB VRAM & Yes & Yes \\
\midrule
Mistral & 7 Billion & Yes & Apache 2.0  & GQA & - & 6GB VRAM & Yes & Yes  \\
\midrule 
Falcon & 7 Billion & Yes & Apache 2.0 & MQA & 1.5 Trillion & 15GB RAM & Yes & Yes \\
Falcon & 40 Billion & Yes & Apache 2.0 & MQA & 1 Trillion & 40GB RAM & Yes & Yes \\
Falcon & 180 Billion & Yes & Apache 2.0 & MQA & 3.5 Trillion & 320GB RAM & Yes & Yes \\
\midrule 
GPT-3 & 175 Billion & Yes & OpenAI License & MHA & 300 Billion & Via API & No & Limited \\
GPT-3.5 turbo & 175 Billion & Yes & OpenAI License & Not disclosed & Not disclosed & Via API & No & Yes \\
GPT-4 & Not disclosed & Yes & OpenAI License & Not disclosed & Not disclosed & Via API & No & No \\
\midrule 
Gemini & 137 Billion & Yes & Gemini Pro License & MQA & Not disclosed & Via API & No & No \\
\midrule 
Claude & 93 Billion & Yes & Claude Pro License & Unknown & Unknown & Via API & No & No \\
Claude 2 & 137 Billion & Yes & Claude Pro License & Unknown & Unknown & Via API & No & No \\
Claude 3 & Unknown & Yes & Claude Pro License & Unknown & Via API & No & No \\
\midrule 
&&&Apache 2.0 for & 48 attention heads &&&& \\
Grok-1 & 314 Billion & Yes & code and  Grok-1  & for queries, & Unspecified & Unspecified & yes & No \\
&&& weights &8 for keys/values&&&& \\
\bottomrule
\end{tabular}
\caption{A comparative summary of the reviewed LLMs\label{tab:LLM-comp}}
\end{table*}
\end{landscape}

\section{Vision Models and Multi-Modal Large Language Models}

In the previous section we introduced the major and most popular of the LLMs originating in the text domain, though some of them have a multi-modal component, we now introduce a specialisation of language models, namely vision models. These are specifically designed to combine text and image data and to create an integrated embedding space unlike those mentioned earlier which have a multi-modal component almost as a retrofit.  Vision models allow for seemless integration of text and imagery in a range of applications including automatic text-to-image generation and image captioning.

\subsection{Vision Models}

\subsubsection{\textbf{\textit{BLIP-2}}}

In June 2023, Salesforce introduced BLIP-2, a pre-training strategy that improves vision-language understanding by utilising pre-trained image encoders and language models. BLIP-2 comprises of two pre-training stages. The first stage uses frozen image encoders to learn visual-text representations while the second stage uses a frozen language model to generate vision-to-language understanding. BLIP-2 introduces a crucial component, the Querying Transformer (Q-Former), which acts as a bridge between image and text encoders. By employing learnable querying vectors, the Q-Former extracts relevant visual features necessary for accurate interpretation of visual data. Comprising both image and text transformers, the Q-Former facilitates interactions between visual features and text. In the first stage of pre-training, the Q-Former learns visual representations most relevant to the text. In the second stage, during the vision-language generative learning, the Q-Former connects to a frozen language model, enabling the model full use of its generative capabilities. The output query embedding from the image transformer is aligned to match with the text embeddings, serving as soft visual prompts that guide the language model’s focus on visual features. This innovative approach enhances vision-language understanding by effectively leveraging pre-trained foundational models while efficiently bridging the modalities~\cite{ BLIP2Breakthrough, li2023blip2}.

\subsubsection{\textbf{\textit{Vision Transformer (ViT)}}}

The Transformer architecture has been shown to be effective in computer vision tasks. In a 2020 paper by Dosovitskiy {\em et al.}~\cite{dosovitskiy2020image}, the authors proposed the Vision Transformer (ViT). They argued that with substantially fewer computational resources, a ViT could be trained with comparable performance to state-of the-art convolutional neural networks, specifically when pre-training on the same computational budget where ViT outperformed classic ResNet architectures~\cite{he2016deep}. Here they reshaped and flattened 2D images into a sequence of patches which can then be utilised by the transformer. In contrast to the transformer for NLP tasks, the Vision Transformer does not have an attention based decoder, and instead passes the output to a multi-layer perceptron head~\cite{dosovitskiy2020image}.

\subsubsection{\textbf{\textit{Contrastive Language–Image Pre-training
(CLIP)}}}

Contrastive Language–Image Pre-training (CLIP) has quickly become the basis for many Multi-Modal Large Language Models (MM-LLMs)
\cite{radford2021learning}. Heavily inspired by the Visual N-Gram approach used by Li {\em et al.}~\cite{li2023evaluating}, it is a multi-modal approach trained on 400,000,000 image-to-text pairs capable of classification without being specifically instructed on the task, i.e. exhibiting zero-shot capabilities. This works by ``jointly training an image encoder and a text encoder" to predict the correct image-text pairings and is capable of matching the performance of ResNet-50 without using its 1.38 million crowd-labelled examples. The potential for encoding bias are addressed here with models like CLIP identified as having particular risk for a developer programming a class into the model. CLIP was assessed for potential degrading and discriminatory classifications by checking the Fairface dataset for non-human classes like `Animal' and `Gorilla' and it was found that CLIP incorrectly classified people of colour for one of these non-human classifications at a disproportionate rate (14\% vs. 8\%). Through further experiments they showed that effective class design could mitigate these problematic classifications.

An improvement to CLIP was recently proposed named Retrieval Augmented Contrastive Language-Image Pre-Training (RA-CLIP)~\cite{10205444} aimed at addressing the data volume requirement of CLIP. Here, the image-text data is sampled as a ``hold-out set" and the gap is then met by online retrieval, with the aim of training the model to recognise visual concepts, and then ``look them up", compared to an open book exam. RA-CLIP claims an improvement on baseline zero-shot image classification of +12.7\%. It is reasonable to assume this method could be used to improve on general purpose vision-language understanding such as LLaVA, as discussed below.

\subsection{Early Approaches to Multi-Modal Information Processing}

Early work on bridging the gap between computer vision and NLP were inspired by machine translation and are known as encoder-decoder methods~\cite{ghandi2023deep}. These converted an image into an intermediate representation before converting to a caption. This typically involved using two neural networks, a Convolutional Neural Network (CNN) for the encoder, followed by using the last layer as the input, the feature space, to a Recurrent Neural Network acting as the decoder.
These types of approaches are computationally expensive and produce vague results. Li {\em et al.}~\cite{li2023evaluating}, showed that images could be annotated in a zero-shot manner by learning from web data, using a similar approach to n-gram language models.

\subsection{Multi-Modal Large Language Models}
Image-grounded MM-LLM's referred to as Large Vision Language Models ``generally consist of vision encoder, a language encoder and a cross-modal alignment network''~\cite{li2023evaluating}. They thus offer a more well-founded pathway to true multi-modal generative AI than the text-based LLMs covered earlier in Section~\ref{sec:LLMs}, some of which have multi-modality added in afterwards. We now explore some of these models.

\subsubsection{\textbf{\textit{Large Language and Vision Assistant (LLaVA)}}}

Large Language and Vision Assistant (LLaVA)~\cite{liu2023visual} introduced a method for connecting a vision encoder and an LLM to produce a MM-LLM for general purpose vision to language understanding. Using CLIP as a visual encoder and Vicuna~\cite{vicuna2023} as the language decoder, the decoder was fine-tuned on 158k language-vision instructions taken from the MS-COCO data set~\cite{lin2014microsoft} with the encoder remaining frozen. The fine-tuning trained on this dataset over 3 epochs with a learning rate of  2e-5 and a batch size of 32. Full Shard Data Parallel  and gradient check pointing was utilised in an effort to save GPU memory. 

This work was expanded by LLaVA 1.5 by adding CLIP-ViT-L-336px with a multi-layer perceptron (MLP) projection. This uses largely the same hyper-parameters as the original LLaVA, excluding the learning rate which has been halved during pre-training.  This is due to the fact the model now uses a MLP projection layer rather than a linear projection layer as in the original model. Another important consideration is that the model can only handle one image at a time~\cite{liu2023improved}.

\subsubsection{\textbf{\textit{Kosmos-1 and Kosmos-2}}}
Kosmos-1 is a MM-LLM claiming to posses general purpose image processing capabilities~\cite{huang2023language}. Kosmos-1 a Transformer-based causal language model. It allows for text as well as other modalities like images to be embedded directly as input to the language model. It converts the different modalities into vectors which can then be passed to the decoder. Kosmos is trained on a combined text corpora of ``The Pile"~\cite{gao2020pile}
and the ``Common Crawl"~\cite{CommonCrawlOrg}, a repository of image-caption pairs, as well as  multi-modal data from the Common Crawl defined as ``Interleaved Image-Text Data". As a common and recurring theme with MM-LLMs in this review, the image representations are given from CLIP. 

Kosmos-2~\cite{Peng2023Kosmos2GM} is built on Kosmos-1 and  has the same Transformer-based language  model architecture and training objective as KOSMOS-1. The training data used as input into the model consists of grounded image-text pairs thus  endowing the model with
grounding and referring capabilities at source. This is an alternative to the approaches taken by the other MM-LLMs described in this paper which train and then combine separate language (text) and vision models in a multi-stage process.

\subsubsection{\textbf{\textit{MiniGPT4}}}

In April 2023, researchers from King Abdullah University of Science and Technology (KAUST) unveiled their findings and access to their  open-source MM-LLM, MiniGPT4. MiniGPT4 was designed to rival the performance of OpenAI’s MM-LLM GPT-4, but as an open-source alternative. Given the lack of transparency regarding the technical design and details behind OpenAI's GPT-4, researchers at KAUST looked to develop a comparable MM-LLM using only open-sourced models and data. The researchers at KAUST hypothesised that the capabilities demonstrated by GPT-4 were attributed to its utilisation of advanced LLMs. To test this hypothesis, MiniGPT4 was developed to examine the use of an open-sourced frozen visual encoder with an advanced open-sourced frozen LLM.  The abilities showcased by MiniGPT4 include the identification of facts and objections from an image, recipe generation from an image, identification of issues with corresponding solutions, and even providing context as to why an image of a meme is considered funny~\cite{zhu2023minigpt4}.

The model architecture of MiniGPT4 utilises an advanced pre-trained LLM as the language decoder called Vicuna. Vicuna was also used in LLaVA as mentioned earlier, and is based on the LLaMA model and fine-tuned using user conversations sourced from ShareGPT.com~\cite{lmsysVicunaOpenSource}. In addition to Vicuna, MiniGPT4 incorporates pre-trained open-sourced vision components similar to those used in BLIP-2. These components include a Q-Former and ViT which serve as the vision encoders within the model architecture. To effectively integrate the pre-trained visual encoder with the LLM, a single projection layer is introduced. This projection layer acts as a bridge, aligning the visual features extracted by the pre-trained visual encoder with the LLM. 

Throughout the pre-training process of MiniGPT4, both the visual encoder and LLM remain frozen, meaning the parameters of these models are not updated. Only the projection layer is trained to align the visual features with the LLM, ensuring effective coordination between the two modalities without altering their individual pre-trained states ~\cite{li2023blip2, zhu2023minigpt4}.

Two stages of fine-tuning were conducted with MiniGPT4 to enhance its performance. Initially, the model was fine-tuned using a dataset comprising approximately  5 million image-text pairs from combined image captioning datasets. This stage of training lasted around 10 hours and utilised 80~GB of A100 GPU. However, the results revealed that training on short image caption pairs led to irregular linguistic and irrelevant content output. It was found that simply aligning visual features with the LLM proved inadequate in producing chatbot quality visual conversation capabilities. Additionally, it was believed that the presence of underlying noise in raw image-text pairs may have contributed to the sub-optimal language outputs. For this reason, and to address this issue, a second fine-tuning stage was implemented. A more detailed image description dataset was created consisting of 3,500 high-quality image-text pairs. The model was further trained using this refined dataset, resulting in MiniGPT4 generating more reliable and natural language outputs. The training time for the refined dataset was reduced to approximately 7 minutes on an A100 GPU. It was found that the additional fine-tuning with more detailed image captions improved the model’s generation reliability and usability.

MiniGPT4's performance was evaluated in comparison to BLIP-2 across four vision-language tasks, each involving presenting the model with an image accompanied by a prompt question. These tasks included meme interpretation, recipe generation, poem composition, and advertisement generation. An example of the prompt question used in the case of meme interpretation task was ``Explain why this meme is funny?”. MiniGPT4 demonstrated successful responses to 65\% of requests, compared to BLIP-2 with less than 10\% success. Additionally, both models were evaluated on image captioning using the MS-COCO caption benchmark~\cite{lin2014microsoft}, with BLIP-2 achieving under 30\% success and MiniGPT4 achieving over 65\%.

Despite MiniGPT4’s notable performance, it was found that the model is subject to the limitations inherent with LLMs, such as hallucinations. The research team at KAUST gauged the hallucination rate of the model's output using the CHAIR metric. They found that longer captions tend to have higher hallucination rates than those of shorter captions and suggested that using reinforcement learning with AI feedback with hallucination detection modules could be a potential solution to the issue.

The research team additionally explored other versions of MiniGPT4, all trained using the same methodology. MiniGPT4 without a Q-Former demonstrated a similar performance to MiniGPT4, although the original version was slightly better. MiniGPT4 with a fine-tuned Q-Former exhibited an even lower performance. Lastly, MiniGPT4 with the incorporated three projection layers performed  poorest, suggesting that the use of a single projection layer is sufficient to align the visual encoder and language decoder~\cite{zhu2023minigpt4}.

\subsubsection{\textbf{\textit{mPLUG-OWL}}}
In April 2023, researchers from the Alibaba DAMO academy introduced mPLUG-OWL, an open-sourced MM-LLM. Other existing MM-LLMs have employed various methods of pre-training and instruction tuning in a two-stage fashion. 

Identifying limitations in these approaches, the DAMO academy researchers focused on addressing constraints associated with relying on frozen visual models throughout both training stages, which could lead to inadequate alignment due to limited parameter flexibility. In response, mPLUG-OWL adopts a novel approach to align image and text, freezing the visual model only in the second stage of training. Additionally, the model leverages a base pre-trained LLM LLaMA as opposed to conversational fine-tuned LLMs such as Vicuna and Alpaca ~\cite{ye2023mplugowl, lmsysVicunaOpenSource}.

The model’s architecture utilises advanced pre-trained LLM LLaMA-7B as the language decoder, while ViT-L/14 serves as the visual foundation model for the visual encoder to extract visual features from the input images, thereby encoding visual knowledge. Recognising the computational complexity associated with integrating visual knowledge into the LLM due to the lengthy sequences, a visual abstractor module is employed to address this issue, by compressing the extracted visual features~\cite{ye2023mplugowl, metaIntroducingLLaMA}. 

The ViT is initialised from the CLIP ViT-L/14 model, which is 
pre-trained to ensure faster convergence. The initialisation allows for the compression of visual information into a few learnable tokens. These compressed visual tokens are then combined with word embeddings from the input sequence, facilitating their integration into the LLM for further processing and analysis. This allows the model to effectively integrate visual information alongside text input, while managing computational resources~\cite{ye2023mplugowl, dosovitskiy2021image}.

The first stage involves pre-training the visual model. This is done by freezing the LLM and training the visual knowledge and abstractor modules on several image-caption pair datasets. In the second stage of training, the LORA fine-tuning technique is implemented on the frozen pre-trained LLM to improve the LLM's ability to efficiently process text instruction data from various sources, while the visual knowledge and abstractor modules are frozen. LoRA is covered in more detail later in Section~\ref{subsec:FT}, on fine-tuning~\cite{ hu2021lora}.
DAMO academy researchers believed that this training approach would enhance multi-modal abilities and improve the generation capabilities of LLMs through modality collaboration by enabling effective integration of textual and visual information. 

The mPLUG-OWL model’s effectiveness was evaluated on OwlEva, a visually related instruction dataset that was created by the DAMO academy researchers that consisted of 82 questions based on 50 images. The model demonstrated promising performance in instruction and visual understanding and knowledge transfer. The model was compared with other MM-LLMs such as LLaVA, BLIP-2 and MiniGPT4, with both MiniGPT4 and mPLUG-OWL showing the most promising results. There were noted incidents of hallucinations in mPLUG-OWL where the model failed to relate some images and produced some text hallucinations~\cite{ye2023mplugowl, dosovitskiy2021image,zhu2023minigpt4, liu2023visual}.

\subsubsection{\textbf{\textit{Re-Cap on Specific MM-LLMs}}}

In comparing the MM-LLMs we have selected for inclusion in this review, we find they are very different in how they have each been constructed in terms of  their interplay between visual and text models.  MiniGPT4 utilised both frozen LLM and frozen visual models in each of two stages, aligning images and text through a projection layer~\cite{zhu2023minigpt4} while LLaVA used a frozen LLM and visual model in the first stage and a trainable LLM with a frozen visual model during the second phase~\cite{liu2023visual, huang2023language}. 
mPLUG-OWL also had a 2-stage approach 
during which the LLM was frozen and the visual model was trained in the first stage, then the LLM was fine-tuned and the visual model was frozen in the second stage of training~\cite{ye2023mplugowl}.  
Kosmos-1 and Kosmos-2 took a different approach and incorporated a trainable LLM alongside frozen visual models in a single combined stage and thus integrated the text and visual training data at input time.

\begin{figure*}[!ht] 
    \begin{center} 
        \includegraphics[width=0.95\textwidth]{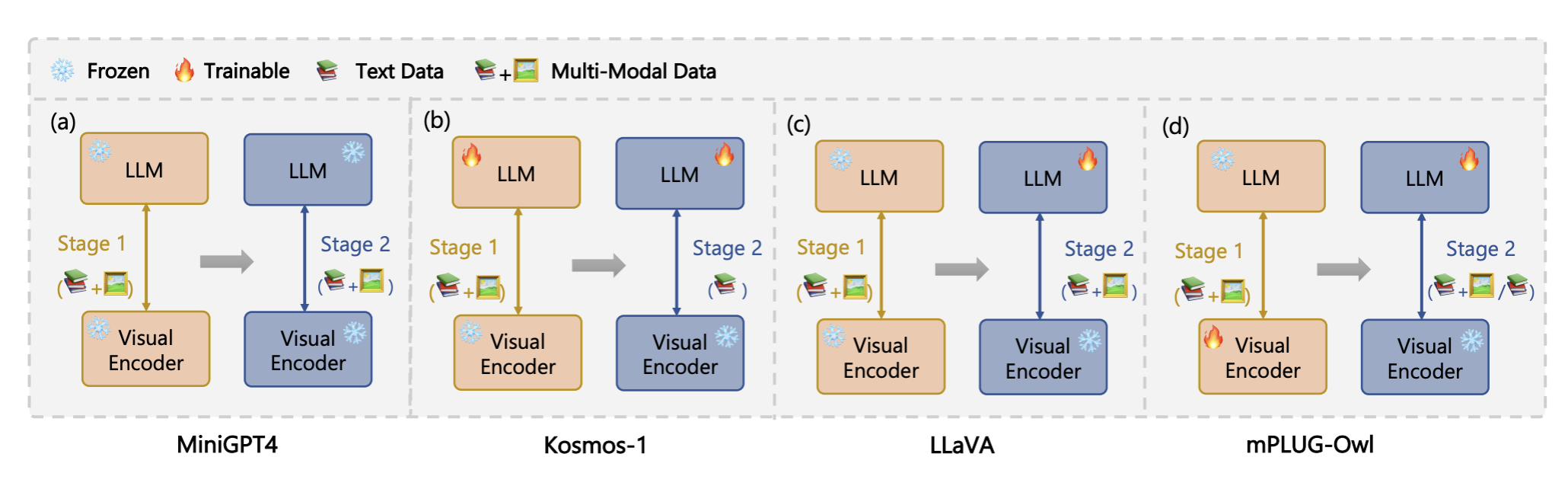}
    \end{center}
    \caption{A comparative summary of different training methods used for the reviewed MM-LLMs, all which follow a two-stage training process (taken from~\cite{ye2023mplugowl}).\label{fig:mplug}}
\end{figure*}

At this time there is clearly no universally agreed way for a foundational text model and a model or training data for other modalities to be co-trained and this question remains an open research issue.
Figure~\ref{fig:mplug} presents an  overview and comparison of the different training methods for MM-LLMs while Table~\ref{tab:MM-LLM-comp} presents a summary of their characteristics.

\begin{table*}[htb]
\centering 
\scriptsize
\begin{tabular}{l|llll}
\toprule
Model	&Open 	&Fine-	&LLM Used&	Vision Model Used \\
& Source & Tuneable &&\\
\midrule 
LLaVA & Yes	& Yes &	Vicuna&	CLIP ViT-L/14 \\
Kosmos-1 and -2 & Yes & Yes & \multicolumn{2}{l}{Grounded image-text pairs to train an integrated model} \\
MiniGPT4 & Yes	&Yes	&Vicuna&	Q-Former and CLIP ViT-G/14 \\
mPLUG-OWL &	Yes &	Yes &	LLaMA-7B &	CLIP ViT-L/14 \\
\bottomrule
\end{tabular}
\caption{A comparative summary of some MM-LLMs\label{tab:MM-LLM-comp}}
\end{table*}

\section{Model Tuning}
There is much potential offered by pre-trained LLMs and MM-LLMs. However, these foundational models may not always align perfectly with the specific requirements of every use case they may be used for. Some scenarios may demand more tailored outputs, while others may lack crucial context or domain-specific knowledge covered in the initial pre-training phase. To fully harness the potential of pre-trained foundational models for specific uses cases, model tuning can be used. There are four primary areas of model tuning that can be utilised to adapt  pre-trained foundational models namely full fine-tuning, parameter-efficient fine-tuning (PEFT), prompt engineering,  and reinforced learning human feedback (RLHF) and these are now described in turn.

\subsection{Full Fine-Tuning}
\label{subsec:FT}
Fine-tuning is a process of modifying some of the parameters of a pre-trained foundational model by further training it on a smaller, task-specific dataset in order to optimise the model’s performance for a target use case.  Full fine-tuning involves adjusting all parameters of a pre-trained model using a task-specific dataset to better align with the specific requirements of the target use case. While this allows for comprehensive customisation, it requires a significant amount of training data. The advantage of full fine-tuning lies in the model’s ability to grasp the nuances of a particular domain. However, this process can be computationally expensive~\cite{deciFullFineTuning, xu2023parameterefficient}. 

\subsection{Parameter Efficient Fine-Tuning}
Unlike full fine-tuning, Parameter Efficient Fine-Tuning (PEFT) aims to reduce computational requirements and training time by selectively updating or modifying only specific parts of a pre-trained foundational model while achieving comparable performance to full fine-tuning. PEFT employs techniques to  refine a pre-trained  model by updating only a small number of its parameters. LLMs and MM-LLMs are pre-trained on extensive datasets that tend to have a wide variety of knowledge, often possessing much of the information required for a range of downstream tasks. Consequently, for many specific targeted use cases, adjusting the entire model may be unnecessary and inefficient. PEFT addresses this by conducting the fine-tuning process on a small subset of parameters, focusing only in those relevant to the specific targeted use case.

There are different PEFT methods employed for fine-tuning to accommodate for different requirements. While certain methods focus on training specific segments of the original model’s parameters, others aim to incorporate and train additional components, such as adapter layers, without altering the original architecture~\cite{deciFullFineTuning, xu2023parameterefficient}. Below we review some of these techniques.

\subsubsection{\textbf{\textit{Low Rank Adaption}}}
Low Rank Adaption (LoRA) trains LLMs or MM-LLMs on domain-specific data and involves freezing the weights of a pre-trained model and injecting trainable rank decomposition matrices into each layer of the Transformer architecture. By utilising this method, it can reduce the number of trainable parameters and GPU memory required for subsequent tasks. LoRA focuses on fine-tuning the smaller trainable matrices that create the LoRA adapter. The LoRA adapter, once fine-tuned, is integrated into the original pre-trained model for inference. The advantage of LoRA is its ability to reduce memory requirements by allowing multiple LoRA adapters to reuse the same original LLM. This makes it useful for handling multiple  tasks and uses cases efficiently~\cite{hu2021lora}.

\subsubsection{\textbf{\textit{Quantised Low-Rank Adaption})} 
Quantised Low-Rank Adaption (QLoRA}
is an extension of the LoRA fine-tuning technique, focusing on memory efficiency. It incorporates quantisation, which involves reducing the precision of numerical values, during the adaption process. 
Instead of fine-tuning all the weights of a pre-trained LLM, the LoRA technique is used to form LoRA adapters which are smaller matrices that can be fine-tuned to approximate the larger weight matrices. However, instead of low-rank approximation, QLoRA also applies quantisation to the weights of the LoRA adapters to further compress the model which reduces the memory and storage requirements even further. Quantisation reduces the precision of numerical data by converting high-precision numerical values to lower precision, resulting in smaller model sizes to preserve performance. In the paper introducing QLoRA, it was found that this fine-tuning approach reduced the memory usage enough to fine-tune a 65B parameter model on a single 48GB GPU~\cite{dettmers2023qlora}.

\subsubsection{\textbf{\textit{Supervised Fine-Tuning}}}
Supervised Fine-Tuning (SFT) involves fine-tuning a domain-specific dataset using labelled training data, on a pre-trained foundational model. LLMs and MM-LLMs are usually trained on extensive unsupervised data, but SFT aims to adapt pre-trained models to learn a domain-specific task from labelled data. By adjusting a model’s weights and parameters to optimise the domain-specific data during fine-tuning, the model becomes specialised in excelling at domain-specific tasks. 

SFT offers a simpler approach for aligning LLMs and MM-LLMs with domain-specific tasks compared to training a model from scratch. The use of SFT can allow developers to gain significant results using less training data and computational resources. However, the resulting models should also be tested for hidden bias in the initial training data used in the pre-trained model. This is to help prevent and avoid hidden biases being magnified when SFT is applied with a domain-specific dataset~\cite{ mediumSupervisedFinetuning, lakeraUltimateGuide}.

\subsection{Prompt Engineering}

Prompts are natural language texts which instruct a LLM or a MM-LLM on the task it should perform. This involves learning a language model that calculates the probability of a given text and uses this probability to predict an output~\cite{10.1145/3560815}. Given such prompts, LLMs are  able to perform ``in-context learning" meaning they have the potential for adaption to a wide variety of tasks without needing new training data~\cite{garg2022can}. This mitigates one of the key problems with fine-tuning a LLM, the need for a large set of training data for each new task~\cite{brown2020language}. 

A distinction can be made between few-shot, one-shot, and zero-shot prompt engineering. Few-shot refers to when multiple examples are given to a model but no weight adjustments are allowed i.e no adjustment to the parameter weights. One-shot as the name suggests is similar to few-shot except that only one example is provided. Zero-shot prompting  is different again in that only the task instruction is given and there is no example provided. 

Arguments have been made that in such cases, LLMs and MM-LLMs are not learning  tasks from few-shot examples but instead performing ``task location in the model's existing space of learned tasks"~\cite{reynolds2021prompt}.
Furthermore, zero-shot performance can equal and even surpass the performance of few-shot prompting. Reynolds {\em et al.} in~\cite{reynolds2021prompt} use the translation task  to show that a small number of samples is insufficient to learn anything about the translation task. They investigate whether any examples are even necessary, to show that the few-shot method is really directing the model to pre-learned information. Sun {\it et al.} argue that while zero-shot performs well, there is a gap when  using the model on domain-specific tasks~\cite{10222039}. To address this, they designed domain-specific prompts from the knowledge base of the LLM and fine-tuned this using a ranking constraint.

\subsection{Reinforced Learning Human Feedback}

Reinforced Learning Human Feedback (RLHF) is a fine-tuning method used to further align the behaviour of foundational models with human preferences. It involves collecting data where human annotators select preferred model outputs, which is then used to train a reward model. The reward model learns patterns in human preferences and guides the LLM or MM-LLM during output generation to produce outputs that align better with human preferences. 

RLHF benefits models by  enhancing their usability in real world applications. The main disadvantage to using RLHF is that it requires a substantial amount of human feedback, training data, computational resources, and there are scalability issues~\cite{touvron2023llama2,mediumWhatRLHF}.

\section{Model Evaluation and Benchmarking}
\label{sec:benchmarks}

A range of benchmarks and evaluation techniques are available for assessing the performance of both pre-trained and fine-tuned LLMs and MM-LLMs. Prior to fine-tuning a pre-trained model, it is important to establish a benchmark performance which then serves as a baseline to measure improvement achieved through fine-tuning. Following the establishment of a baseline, a model’s performance post fine-tuning can be evaluated to assess the effectiveness of the fine-tuning process. For  evaluating domain-specific fine-tuned models, specialised exams in the respective domain have been used or created. For example, in the case of MedAlpaca, a medically fine-tuned LLM, the domain-specific fine-tuned model was assessed using a specialised medical exam, the USMLE, a standardised assessment taken by medical students to qualify as physicians~\cite{han2023medalpaca}. The model’s performance was evaluated based on the answers produced with zero-shot prompts and performance was compared with other models for benchmarking purposes.

In addition to accuracy, it is important to assess a pre-trained foundational model prior to fine-tuning for issues in areas such as hidden biases, common sense reasoning, discrimination, and contextual understanding. To evaluate LMMs for toxic language like hate speech or threats, the RealToxicityPrompts benchmark~\cite{gehman2020realtoxicityprompts} has been utilised. A model completes 100k prompts in which a toxicity score is evaluated through a request to the Perspective API~\cite{10.1145/3534678.3539147}. 
For reviewing LMMs for biases across various areas such as gender, religion, race, to physical appearance and socioeconomic status, the CrowS-Pairs benchmark can be employed. This benchmark assesses a model’s response and perplexity to examples created using both a stereotype and an anti-stereotype, prior to any fine-tuning or training on those examples~\cite{touvron2023llama, mediumReviewLLaMAOpen}.

Another benchmark used to assess bias in gender is the WinoGender social bias benchmark. It evaluates a model’s capability to resolve co-references influenced by the gender of the pronouns, gauging its ability to accurately link pronouns to previously mentioned nouns. Consistent misinterpretations of gender pronouns may indicate biases or limitations in the model, thus allowing for the evaluation of bias levels. The Winogrande benchmark evaluates a models’ contextual understanding by presenting two similar sentences with a trigger word, where the correct answer depends on comprehending the context~\cite{whytryaiBenchmarksWhat, touvron2023llama, mediumReviewLLaMAOpen}.

For evaluating a models’ ability to handle common-sense reasoning, several popular benchmarks exist, including the following:
    \begin{enumerate}
        \item AI2 Reasoning Challenge (ARC): This test assesses knowledge and common-sense reasoning through grade-school level multiple choice questions~\cite{clark2018think}.
        \item HellaSwag: This task evaluates common-sense reasoning by requiring models to complete sentences based on everyday events, evaluating natural language inference~\cite{zellers2019hellaswag}.
        \item BoolQ: This benchmark consists of real yes/no questions from Google searches paired with Wikipedia passages. It challenges models to infer answers from context that may be implied but not stated~\cite{clark2019boolq}.
        \item OpenBookQA: This question-answering dataset modelled after open book exams used for assessing human understanding of various subjects~\cite{mihaylov2018can}.
        \item PIQA (Physical Interaction Question Answering): This benchmark evaluates a models’ knowledge and understanding of the physical world by presenting hypothetical scenarios with specific goals and multiple choice solutions~\cite{bisk2020piqa}.
        \item Multitask Language Understanding (MMLU): This benchmark measures LLM knowledge across multiple different subject areas using multiple choice questions~\cite{hendrycks2020measuring}.
    \end{enumerate}

\noindent
In order to gauge the accuracy of a foundational model’s outputs concerning misinformation or hallucinations, various other benchmarks have been utilised. One such benchmark is TruthfulQA~\cite{lin2022truthfulqa}, which is designed to assess the truthfulness of a model’s responses. 
It achieves this by querying a model's responses on 817 questions of various styles  across a range of 38 diverse categories, intentionally constructed to challenge the model’s comprehension and accuracy. The output generated by the model is then scrutinised for any signs of misinformation. Another benchmark, M-HALDetect~\cite{gunjal2023detecting}, serves as a dataset specifically tailored for evaluating a model’s tendency for object hallucinations. This benchmark is important for identifying instances where the model may generate outputs containing false or misleading information~\cite{ BenchmarksIntroduction, whytryaiBenchmarksWhat, touvron2023llama, mediumReviewLLaMAOpen}.

While these evaluations and benchmarks are important, they can only be used as an indicator of potential risks or issues within LLMs or MM-LLMs. However, they are not adequate enough to ensure that there is absolutely no risks or issues remaining within the model.

\section{Conclusions}
Over the past few years there has been a sea-change in how NLP is  used to address challenges and provide real world services. With the Transformer architecture and by extension LLMs, we now see sophisticated conversational AI tools which show impressive reasoning and problem-solving skills. This has led to a crossover into the field of computer vision to a point where we are now seeing so-called MM-LLMs and LVLMs emerge such as LLaVA and mPLUG-OWL where vision models are combined with a language-based LLM. By fine-tuning a model or performing prompt-engineering such models have been shown to be adaptable to new tasks in domain-specific areas. However, these models are not without their problems, including the tendency to hallucinate. It is not yet possible to completely eliminate hallucinations, through techniques like Visual Contrastive Decoding they can be reduced. 

There are benefits to utilising open-sourced LLMs and MM-LLMs, with researchers and developers having complete control over their training data, which is an important consideration when handling  sensitive information. Additionally, from assessing the technical aspects of open-source MM-LLMs, they do not always use the largest LLM available. For instance, models such as MiniGPT4 and mPLUG-OWL use LLaMA-7B and Vicuna-13B respectively instead of newer models like LLaMA-2, potentially due to the computational expense of employing techniques such as RLHF. Moreover, LLMs like Falcon demonstrated that the use of high-quality data during pre-training can significantly enhance the downstream performance of LLMs, highlighting the importance of considering the quality of data being used when fine-tuning a MM-LLM.

It can be seen that a range of factors influence the performance and usability of MM-LLMs. Architectural components affect computational resourcing. Attention mechanisms can improve performance but may reduce the detail retained compared to other mechanisms. Additionally implementing fine-tuning methods does not always guarantee better results, as demonstrated with MedAlpaca-13B LoRA versus MedAlpaca-13B, while some fine-tuning methods may be too computationally expensive to employ. 

Although there are benchmarks that can be harnessed to gauge if there are issues or risks within a model, there is no guarantee that the model is completely free of inheriting certain risks or issues.
Finally, in order to evaluate the quality of a fine-tuned domain-specific MM-LLM, there is a need to  consider  how it will be assessed and evaluated. As demonstrated with MedAlpaca, that model utilised the USMLE assessment to test domain relevance overall, and navigating these factors is crucial for developing and evaluating a high-quality MM-LLM for domain-specific applications.

\printbibliography

\end{document}